\documentclass[letterpaper, 10 pt, conference]{ieeeconf}  

\IEEEoverridecommandlockouts                              

\usepackage{graphicx}
\usepackage{hyperref}
\usepackage{caption}
\usepackage{subcaption}
\usepackage{multirow}
\usepackage[T1]{fontenc}
\usepackage{array}
\usepackage{float}
\usepackage{textcomp}
\usepackage{amsmath}
\usepackage{amssymb}
\usepackage{stackrel}
\usepackage{lipsum}
\usepackage{booktabs}
\usepackage{url}
\usepackage{threeparttable}
\usepackage{algorithm}  
\usepackage{algpseudocode}  
\title{\LARGE \bf
Learning-based Optoelectronically Innervated Tactile Finger for Rigid-Soft Interactive Grasping
}
\author{Linhan Yang$^{1}$, Xudong Han$^{2}$, Weijie Guo$^{2}$, Fang Wan$^{3}$, Jia Pan$^{4}$, and Chaoyang Song$^{5,*}$
    \thanks{*This work was supported by National Science Foundation of China (No. 51905252), Shenzhen Institute of Artificial Intelligence and Robotics for Society Open Project (No. AC01202005003), Shenzhen Long-term Support for General Project, Guangdong Provincial Key Laboratory of Human-Augmentation and Rehabilitation Robotics in Universities, Department of Mechanical and Energy Engineering, Southern University of Science and Technology, 518055, China, and AncoraSpring Inc.}
\thanks{$^{1}$Linhan Yang is a Ph.D. candidate with University of Hong Kong, Hong Kong 999077 and Southern University of Science and Technology,
        Shenzhen, Guangdong 518055, China.
        {\tt\small yanglh9652@gmail.com}}%
\thanks{$^{2}$Xudong Han and Weijie Guo are with the Department of Mechanical and Energy Engineering, Southern University of Science and Technology, 
        Shenzhen, Guangdong 518055, China. 
        {\tt\small \{11812519, 11510615\}@mail.sustech.edu.cn}}%
\thanks{$^{3}$Fang Wan is with SUSTech Institute of Intelligent Manufacturing, Southern University of Science and Technology, 
        Shenzhen, Guangdong 518055, China. 
        {\tt\small sophie.fwan@hotmail.com}}%
\thanks{$^{4}$Jia Pan is with Department of Computer Science, University of Hong Kong, 
        Hong Kong 999077. 
        {\tt\small jpan@cs.hku.hk}}
\thanks{$^{5}$Chaoyang Song is the corresponding author with the Department of Mechanical and Energy Engineering, Southern University of Science and Technology,
        Shenzhen, Guangdong 518055, China.
        {\tt\small songcy@ieee.org}}%
}

\begin{document}
\maketitle
\thispagestyle{empty}
\pagestyle{empty}
\begin{abstract}
    
    This paper presents a novel design of a soft tactile finger with omni-directional adaptation using multi-channel optical fibers for rigid-soft interactive grasping. Machine learning methods are used to train a model for real-time prediction of force, torque, and contact using the tactile data collected. We further integrated such fingers in a reconfigurable gripper design with three fingers so that the finger arrangement can be actively adjusted in real-time based on the tactile data collected during grasping, achieving the process of rigid-soft interactive grasping. Detailed sensor calibration and experimental results are also included to further validate the proposed design for enhanced grasping robustness.
    
\end{abstract}
\section{Introduction}
\label{sec:Introduction}

    Data-driven grasp learning has been a research field of growing interest in the past decade \cite{bohg2013data} with many literature contributing to the use of computer vision for grasp prediction \cite{pinto2016supersizing, lenz2015deep, mahler2019learning}, high-resolution tactile sensing \cite{yuan2017gelsight, yamaguchi2016combining}, and advanced gripper design \cite{ma2017yale, yuan2020design}. Many dataset have been published to support grasp learning using computer vision \cite{lenz2015deep, chu2018real}, model-based synthetic methods \cite{mahler2019learning, mahler2017dex}, or direct grasp trials \cite{pinto2016supersizing, levine2018learning}. However, vision is not the only part of the stimuli humans rely on while manipulating objects. Tactile information plays an important role when humans are executing dexterous manipulation, especially when visual information is not precise or fully available \cite{fazeli2019see}. Recent research on using vision sensors for tactile sensing \cite{yuan2017gelsight, yamaguchi2016combining} leverages material properties such as transparency and softness to capture refined detail of physical interaction. Proprioception is another way to collect tactile information to infer data of physical interaction between the finger and the objects, where soft finger with compliant materials is often considered as a feasible choice of design \cite{van2018soft, thuruthel2019soft}.

\begin{figure}[t]
        \begin{centering}
        \textsf{\includegraphics[width=1\columnwidth]{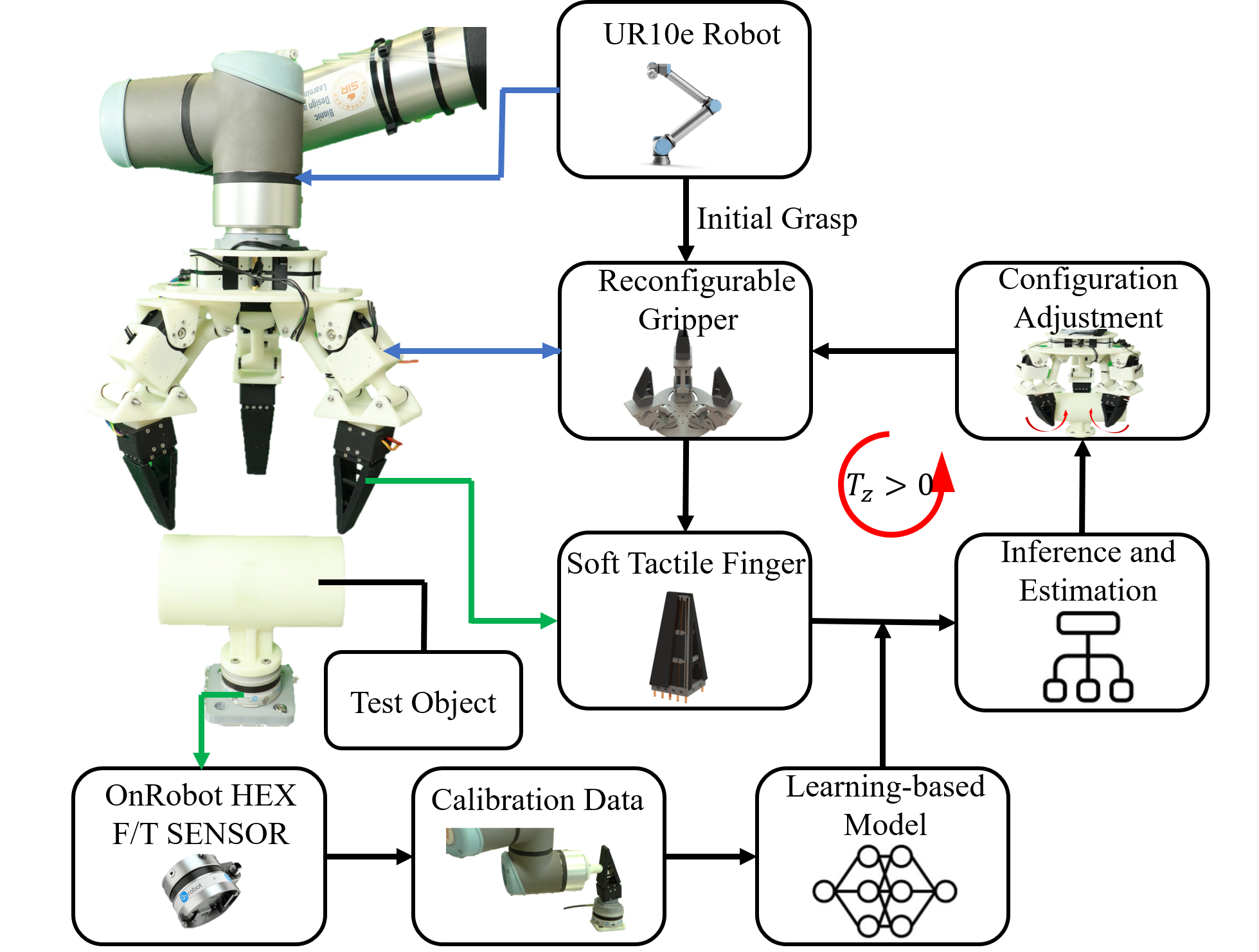} }
        \par\end{centering}
        \caption{Overview of the proposed system design with omni-adaptive tactile fingers for rigid-soft interactive grasping learning. Left shows the system integration of a reconfigurable gripper with the proposed tactile fingers on a UR10e robot. Right shows the workflow of the proposed rigid-soft interactive grasping policy.}
        \label{fig:PaperOverview}
    \end{figure}

    This paper proposes a real-time policy for robust grasping, as shown in Fig. \ref{fig:PaperOverview}, using tactile data collected from a novel design of optoelectronically innervated tactile finger for the rigid-soft interaction between rigid objects and soft fingers.

\subsection{Soft Robotic Fingers and Grippers}

    Robotic fingers and grippers with a soft design have shown great potentials in grasping \cite{shintake2018soft}. By leveraging material softness, they usually feature passive compliance \cite{crooks2017passive, yang2020rigid, wan2020reconfigurable} and underactuation \cite{ma2017yale, odhner2014compliant, deimel2016novel}, leading to a simple control during grasping. Inspired by the Fin-Ray Effect, \cite{crooks2017passive} proposed a novel compliant lattice-structure fingers to enhance grasp robustness and has been widely used in \cite{crooks2017passive, hashizume2019capacitive, basson2019geometric}. Most of these grippers have only one actuator, and the adaptive nature of such finger design can quickly achieve form closure of the target object. Grippers with reconfigurable finger patterns is another solution to enhance grasping robustness. Dollar et al. \cite{ma2017yale, odhner2014compliant} proposed a series of the under-actuated grippers, which could reconfigure between power-grasping, spherical-grasping, and lateral-grasping modes. However, all the control algorithms of different grasping models are open-loop and hard-coded, which may not be applicable for unstructured interaction.

\subsection{Proprioceptive Sensing}

    Stretchable sensors such as soft resistive and soft optoelectronic ones have been embedded in soft robotic fingers to sense the interaction \cite{van2018soft, thuruthel2019soft, zhao2016optoelectronically, stupar2012wearable, zimmerman1985optical}. Zhao et al. \cite{zhao2016optoelectronically} proposed a prosthetic hand using optical waveguides for strain sensing, which could be used to sense the shape and softness of selected objects. Meerbeek et al. \cite{van2018soft} presented an elastomeric foam with 30 optical fibers internally illuminated. They trained this foam sensor system with a machine learning method to detect bending and twisting. Thuruthel et al. \cite{thuruthel2019soft} proposed a human-inspired soft pneumatic actuator with embedded soft resistive sensors for real-time modeling of its kinematics using recurrent neural networks. A review of the above research shows the great potentials of the soft fingers and grippers with proprioception. However, it remains a challenge to apply these novel designs to robotic grasping or dexterous manipulation. 

\subsection{Proposed Method and Contributions}

    This paper is a continuation of our earlier work with the omni-adaptive soft finger for rigid-soft interaction learning \cite{yang2020rigid}, finger configuration learning \cite{wan2020reconfigurable}, and optical fiber-based grasp sensing \cite{Yang2020Scalable}. In this paper, we propose a sensorized design of the omni-adaptive soft finger using multiple optical fibers embedded with friction enhanced soft surface. Tactile data such as normal force, torque, and contact position can be learned based on the proposed design, which is further integrated into a reconfigurable gripper to achieve rigid-soft interactive grasping with enhanced robustness. The contributions of this paper are listed as the following: 
    \begin{enumerate}
        \item Proposed an integrated design of the omni-adaptive soft finger with enhanced finger surface and multi-channel optical fiber for proprioceptive, tactile sensing.
        \item Investigated a detailed characterization and calibration of the tactile finger's sensing capability using machine learning.
        \item Achieved real-time sensing of multiple tactile data, such as normal force, torque, and contact position, with the proposed finger design.
        \item Integrated the proposed tactile finger in a reconfigurable, three-finger gripper that could grasp and adapt to the target object online based on real-time tactile data.
    \end{enumerate}

    In the rest of this paper, Section \ref{sec:Method} explains the problem formulation of rigid-soft interactive grasping and proposed method. Section \ref{sec:Calibration} conducts sensor calibration through machine learning. Experimental results and discussion are enclosed in sections \ref{sec:Experiment} and \ref{sec:Discussion}. Final remarks are enclosed in section \ref{sec:Conclusion}.

\section{Method}
\label{sec:Method}
\subsection{Problem Formulation}

    The research problem of interest is investigating the potentials of tactile sensing for robust grasp learning by leveraging soft robot designs with optoelectronic sensing. We organize this research by building upon our earlier research on a soft robotic finger design with omni-directional adaptation \cite{yang2020rigid, wan2020reconfigurable, Yang2020Scalable}. 
    
    \begin{itemize}
        \item We first discovered the potentials of rigid-soft interactive learning with such finger design, where grasp training could be achieved using much less data when data of soft finger picking up rigid objects are fused in the training dataset \cite{yang2020rigid}.
        \item We then explored the optimal number and arrangement of soft fingers for robotic gripper using a learning-based method and found that three-finger configuration with a centric arrangement is among best ones \cite{wan2020reconfigurable}.
        \item Recently, we further improved the soft finger design by embedding optical fibers inside to enable scalable tactile sensing for object classification during grasping \cite{Yang2020Scalable}.
    \end{itemize}

    In this paper, our goal is to optimize grasp quality by adjusting the finger configuration online based on the real-time tactile information collected from the soft fingers. The contact between the gripper and the object during interaction is assumed as planar. We simplified this problem by discretizing the contact plane into several or a single point. The single finger contact model and multi-finger grasping model are discussed, respectively. 
    
\subsubsection{Contact model of a single finger}
    
    We assume that 1) the grasped object is light such that its weight is neglected in the balancing equation, and 2) the torques along the X- and Y-axis are neglected for their little influence on the overall grasping performance. 
    
    \begin{figure}[htbp]
        \begin{centering}
        \textsf{\includegraphics[width=1\columnwidth]{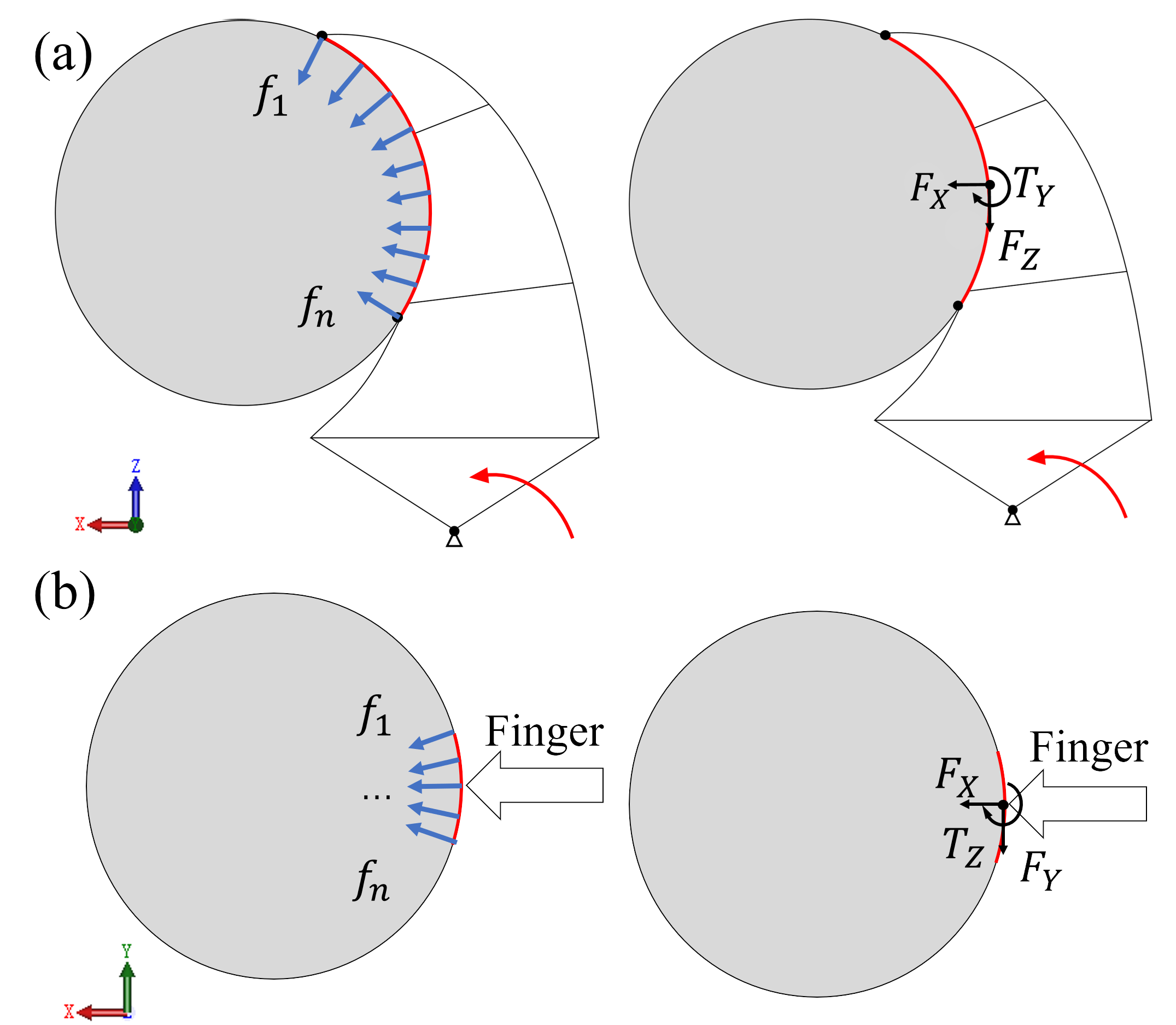} }
        \par\end{centering}
        \caption{Schematic of a single finger contact in (a) X-Z cross section and (b) X-Y cross section.}
        \label{fig:SingleFingerContact}
    \end{figure}

    In the simplified single finger case illustrated in Fig. \ref{fig:SingleFingerContact}, the contact force is discretized as $\boldsymbol{F} = (f_{1}, f_{2}, ..., f_{n})$. In the X-Z cross section shown in Fig. \ref{fig:SingleFingerContact}(a), the resultant force of contact is capable of transmitting a normal force component $F_{x}>0$, a frictional force component $F_{z}$ and a balancing torque component $T_{y}$ along Y-axis, which is neglected under our assumptions. In the X-Y cross section shown in Fig. \ref{fig:SingleFingerContact}(b), the resultant contact force is capable of transmitting a normal force component $F_{x}>0$, a frictional force $F_{y}$ and a balancing torque component $T_{z}$ along Z-axis.
    
    According to Coulomb friction model, when object is grasped, we have: 
    \begin{equation}    \label{eq:Coulomb friction}
    	|| F_{t} || \leq \mu F_{n}, v = 0
    \end{equation}
    where $\mu$ is the friction coefficient and $v$ is the sliding velocity of object. If the sliding velocity is zero, then the magnitude of the tangential friction force is less than or equal to $\mu$ times the normal force, which is non-negative. As shown in Fig. \ref{fig:SingleFingerContact}, the normal force $F_{n} = F_{x}$ and the tangential frictional force $F_{t} = F_{y}+ F_{z}$. Then we get:
    \begin{equation}    \label{eq:Coulomb friction2}
    	|| F_{y} + F_{z} || \leq \mu F_{x}, v = 0
    \end{equation}

\subsubsection{Grasp model of multiple fingers}

\begin{figure}[htbp]
        \begin{centering}
        \textsf{\includegraphics[width=1\columnwidth]{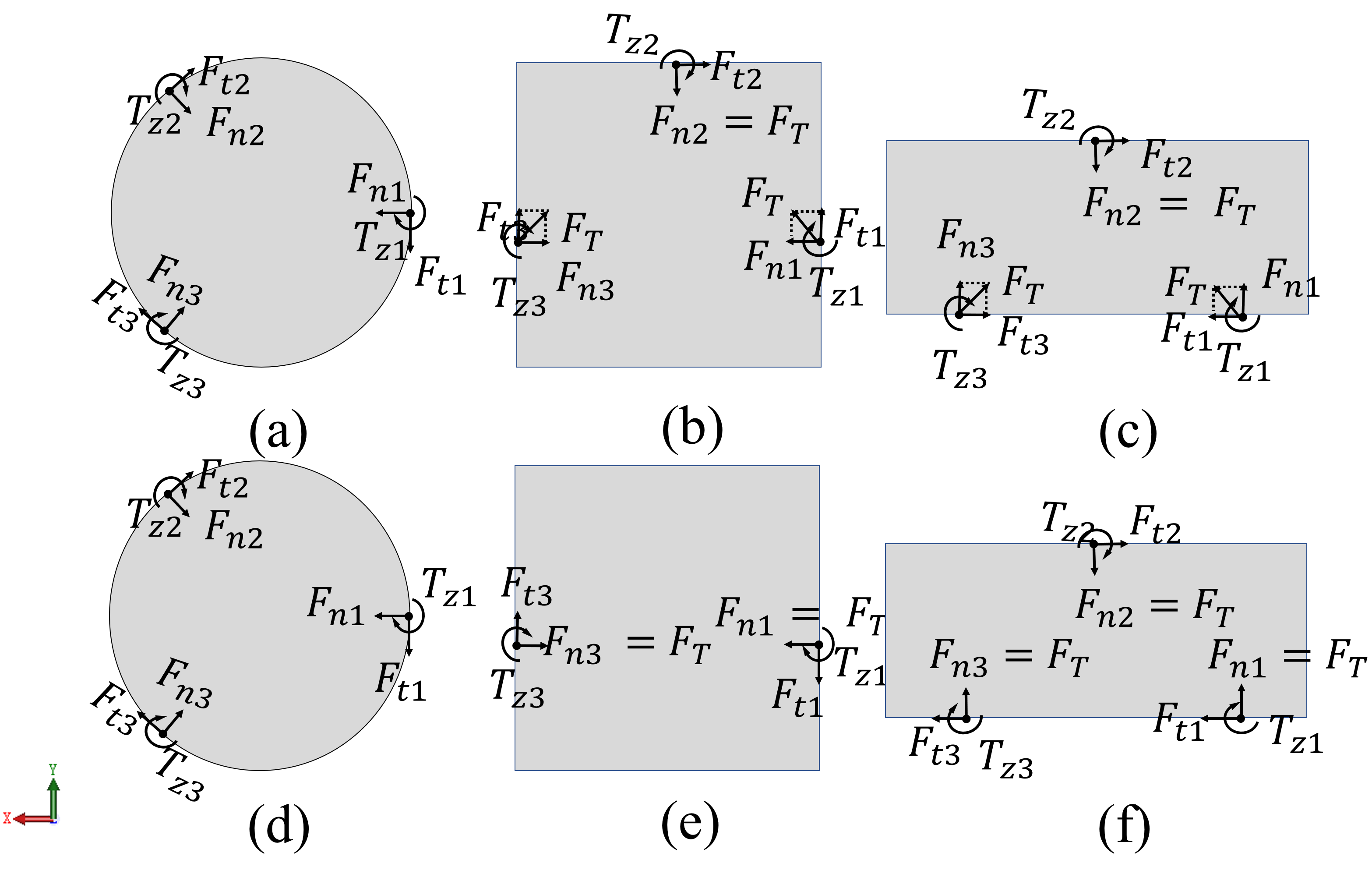} }
        \par\end{centering}
        \caption{Schematic of grasp model. Upper: Conventional configuration. All with circular configuration; Bottom: Interactive grasping which transition between different configuration to conform the surface of objects (d) Circular Configuration; (e) Lateral Configuration; (f) Parallel Configuration. Note: $F_{T}$ denote force provided by actuator  which is a constant in this paper.}
        \label{fig:MultiFingerGrasp}
    \end{figure}

    Based on our earlier research \cite{wan2020reconfigurable}, different number and arrangement of fingers significantly influence the grasping outcome and robustness of the gripper. For example, when the gripper has three fingers, we can model its grasping with three contact areas, with one for each finger, as illustrated in Fig. \ref{fig:MultiFingerGrasp}. 
    
    According to the Newtonian mechanics equilibrium equation, when sliding velocity $v = 0$, we have:
    \begin{equation}    \label{eq:Nowtonian mechanics1}
    	\stackrel[i]{}{\sum}F_{n,i} + \stackrel[i]{}{\sum}F_{t,i} + F_{ext} = 0,
    \end{equation}
    and
    \begin{equation}    \label{eq:Nowtonian mechanics2}
    	\stackrel[i]{}{\sum}T_{z,i} + T_{ext} = 0 \; (i = 1, 2, 3)
    \end{equation}

    From Eq. (\ref{eq:Coulomb friction}), we have
    \begin{equation}    \label{eq:Nowtonian mechanics3}
    	|| F_{t,i} || \leq \mu F_{n, i} \; (i = 1, 2, 3)
    \end{equation}

\subsection{Optoelectronically Innervated Soft Tactile Finger}
    
\begin{figure}[htbp]
        \begin{centering}
        \textsf{\includegraphics[width=1\columnwidth]{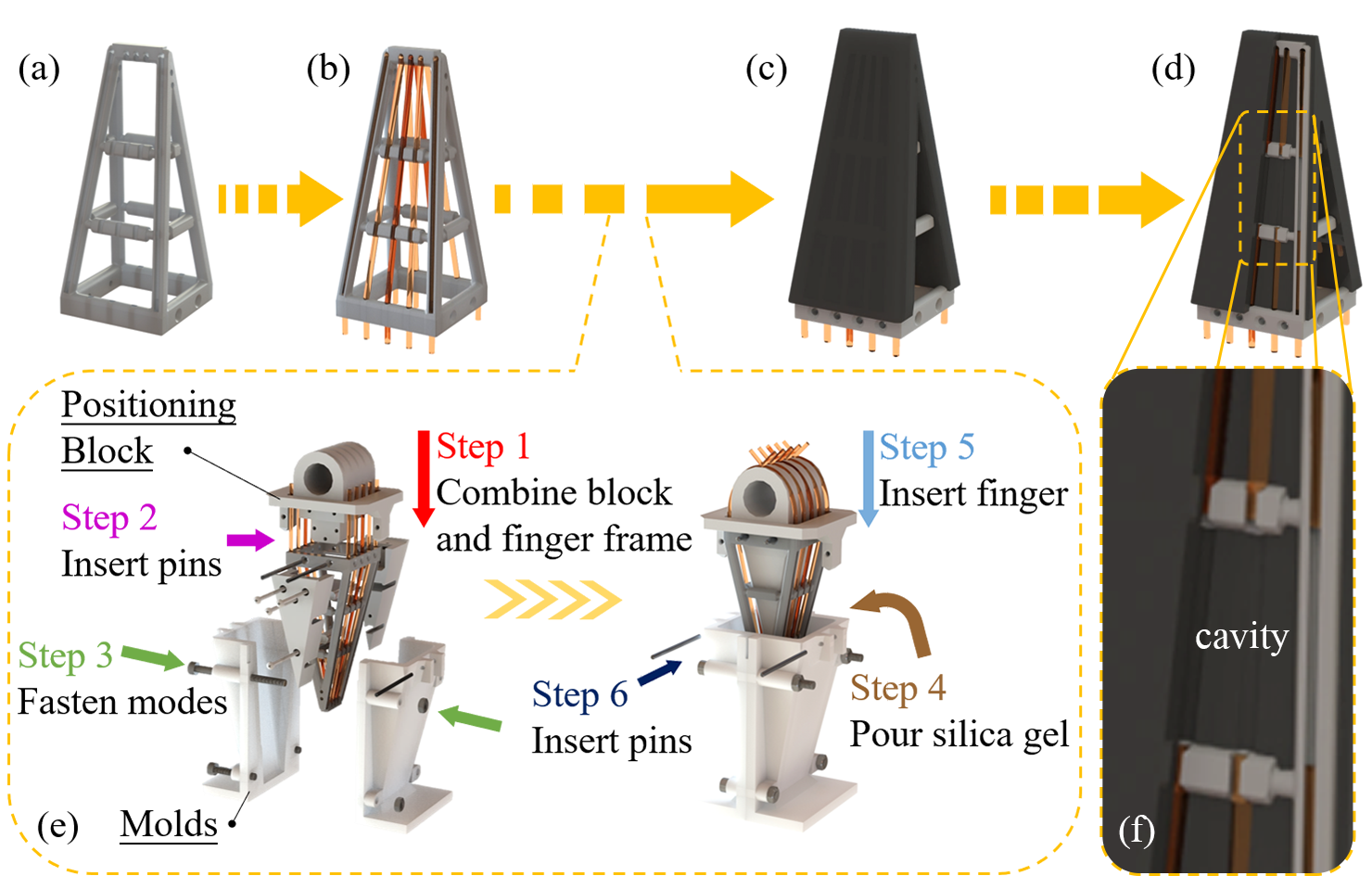}}
        \par\end{centering}
        \caption {The design and fabrication process of the optoelectronically innervated soft tactile finger: (a) finger frame; (b) finger frame with the fibers (orange transparent material is used instead to visualize the transparent fibers clearly in the figures); (c) finger frame with silica gel skin; (d) we pull out the optic fiber to leave a cavity in the middle of finger; (e) fabrication process of the black silica gel skin; (f) the cavity in the sensitive area.}
        \label{fig:FingerDesign}
    \end{figure}

    In our recent work \cite{Yang2020Scalable}, we leveraged the structural space within our earlier finger design \cite{yang2020rigid, wan2020reconfigurable} by introducing optical fibers inside the soft structure to achieve tactile sensing for a limited object classification. In this paper, we systematically improved the finger design, as shown in Fig. \ref{fig:FingerDesign}, with much enhanced sensor reliability, tactile capability, and grasping robustness. Major enhancement include 
    \begin{itemize}
        \item Added finger surface for enhanced contact friction while preserving its omni-directional adaptation;
        \item Multi-channel optical fingers on the main side of grasping with an integrated design and enhanced tactile sensing;
        \item Enhanced fabrication process for improved reliability and consistence performance with reduced cost and complexity.
        \item Maintained compatibility with unstructured environmental changes as before, even with the enhanced design features.
    \end{itemize}
    
    In this new finger design, five optical fibers form a sensor array inside the newly introduced finger surface and the soft finger structure to measure the finger deformation during grasping, as shown in Fig. \ref{fig:FingerDesign}(b). The luminous flux loss correlates to the soft finger's curvature, which makes the optical fiber a feasible choice to obtain the soft finger's deformation. Five LEDs are placed at the transmitting ends of the optical fibers as the luminous transmitter, and five photoresistors are placed at the receiving ends to estimate the output luminous intensity. To reduce the ambient light's impact and get enough luminous flux, we chose 520-525NM LEDs and 520-550NM photoresistors, whose wavelength segments are more concentrated and matching. The result is a significant increase in the signal-to-noise ratio for the required sensing. 
    
    Moreover, the array of optical fibers were placed on the finger framework and covered with a layer of skin made of black silica gel for isolation from the environment, as shown in Fig. \ref{fig:FingerDesign}(c), contributing to the increased signal-to-noise ratio. We further left a cavity in the middle segment of the contacting surface where the major deformation occurs to improve the optical fiber's sensitivity \cite{stupar2012wearable}. As shown in Fig. \ref{fig:FingerDesign}(d)(f), each one of the five optical fibers was cut into two segments and is discontinued at the cavity segment highlighted in Fig. \ref{fig:FingerDesign}(f).

    The fibers we used are commercial optical fibers of soft PMMA manufactured by EverhengFiber$^\circledR$. The skin is made of Smooth-On Ecoflex$^\text{TM}$ 00-30 silica gel, whose strength is suitable for our purpose. However, this silica gel is originally a milky white translucent liquid, which is later mixed with black pigment at a ratio of 20:1 to block the ambient light effectively. Moreover, the silica gel skin is set 3mm away from the finger's outer surface to increase the finger surface texture and enhance the grasping effect. Fig. \ref{fig:FingerDesign}(e) shows the details of the black silica gel skin's fabrication process.

    Let $I_{0}$ denotes the baseline luminous intensity without any deformation. With the current output luminous intensity $I$, the luminous flux loss in decibels through the optical fiber is then described as
    \begin{equation}    \label{eq:luminous flux loss}
         a = 10 \log_{10}(I_{0}/I)
    \end{equation}
    By this definition, the output loss $a$ is 0 without deformation and less than 0 when interacting with the environment.
    
\subsection{Reconfigurable Gripper Design of Three Fingers}

\begin{figure}[htbp]
        \begin{centering}
        \textsf{\includegraphics[width=1\columnwidth]{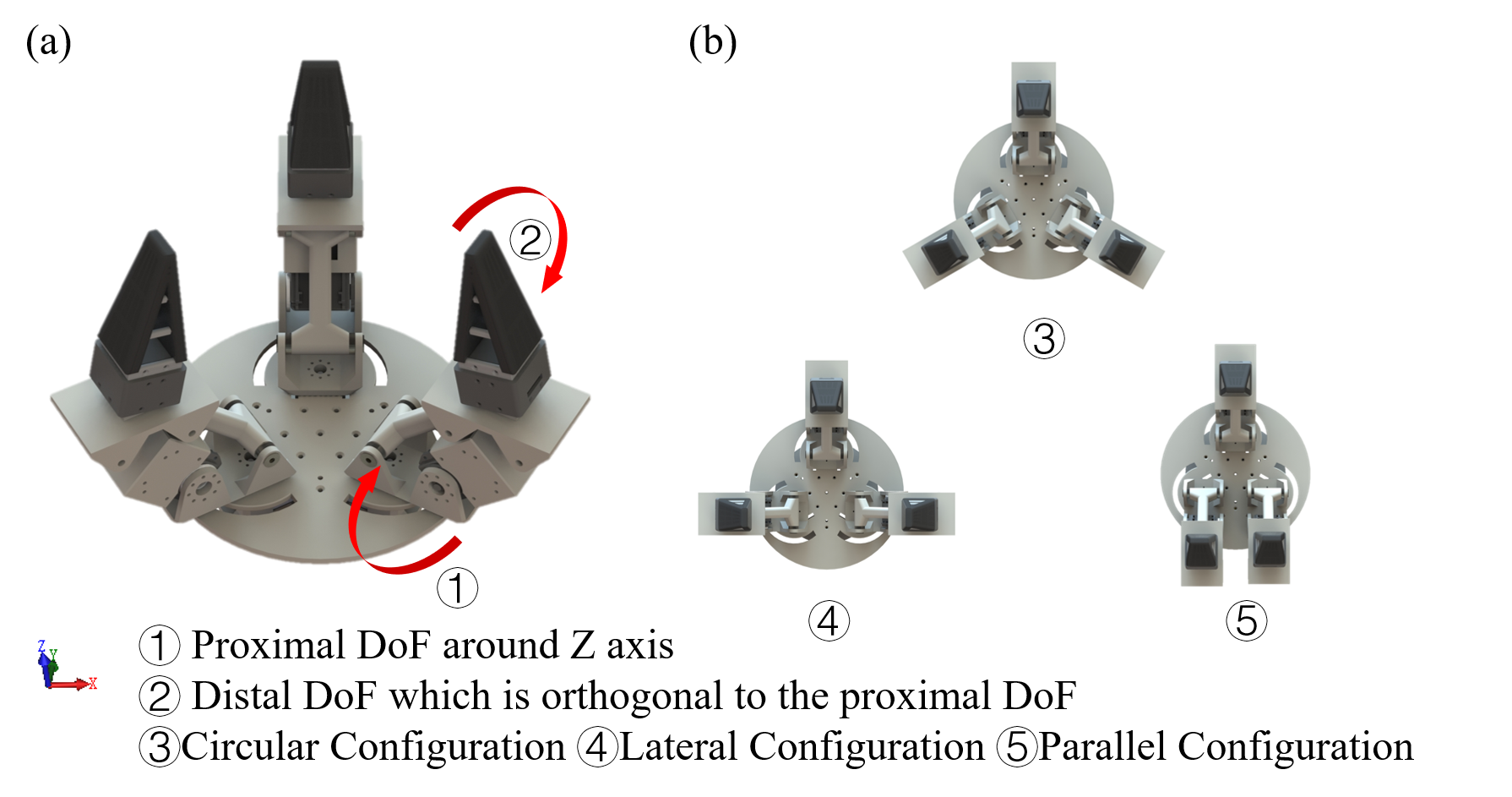}}
        \par\end{centering}
        \caption{CAD model of gripper: (a) three fingers each with two degrees of freedom, (b) three different configurations for grasping.}
        \label{fig:GripperDesign}
    \end{figure}

    Building on our earlier results regarding the optimal number and arrangement of fingers for robotic grippers \cite{wan2020reconfigurable}, we designed a three-finger gripper for this study, as shown in Fig. \ref{fig:GripperDesign}(a). Each finger has two Degree-of-Freedoms (DoFs), with one (ROBOTIS Dynamixel XM320) actuating the finger's opening and closure, and the other (ROBOTIS Dynamixel MX64AT) redirects the fingers' orientation about the Z-axis. Both actuators feature a fully integrated motor with relatively low price in a compact form factor and are daisy-chained to a 12V power supply. High-level Python API provided by Dynamixel SDK is used for communication between the gripper hardware and host computer, providing information on the current position, the goal position, and the torque limit for adjusting the finger's stiffness.

    As shown in Fig. \ref{fig:GripperDesign}(b), this gripper exhibits three-finger configurations: 1) Circular Configuration (i.e., three fingers circularly arranged, facilitating the grasp of a spherical object); 2) Lateral Configuration (i.e., two fingers symmetrically arrange with the third one placed such that the proximal joint axis is on the mid-plane of the others, facilitating more precise grasping); and 3) Parallel Configuration (two fingers on the same side and parallel to the third finger on the other side, facilitating the grasp of cylindrical objects). Such a gripper can adapt to various geometric features of the target objects using the three typical configurations.

\subsection{Rigid-Soft Interactive Grasping}

    The objective is to find a robust grasp policy that maintains the equilibrium equation under external disturbances. We assume a constant $F_{n, i}$ with no external disturbance, 
    \begin{equation}    \label{eq:equilibrium equation with no disturbance}
        \stackrel[i]{}{\sum}F_{t_{0},i} = -\stackrel[i]{}{\sum}F_{n_{0},i}
    \end{equation}

    Then, the anti-disturbance ability could be defined as the minimum of $F_{ext}$, which could break the equilibrium state:
    \begin{equation}    \label{eq:min Fext equation}
        F_{ext} =  \max \stackrel[i]{}{\sum}F_{t,i} - \stackrel[i]{}{\sum}F_{t_{0},i}
    \end{equation}
    where $\max F_{t,i} = \mu F_{n,i}$. To maximize anti-disturbance ability, we should minimize $ F_{t_{0},i}$. and maximize $ F_{n,i}$. 
    
    As shown in Fig. \ref{fig:MultiFingerGrasp}, when grasping with a conventional configuration, 
    \begin{equation}    \label{eq:force isolation}
        F_{n} = F_{T} \cos \theta,
    \end{equation}
    where $F_{T}$ is the constant force provided by actuator and $\theta$ is the angle between $F_{T}$ and the normal direction. During this process, the soft finger is usually twisted and $F_{T}$ is decomposed, which causing a smaller $F_{n}$ and a larger $F_{t}$. 
    
    Grasping with an interactive configuration actively adapted to the surface of the target objects. During the adjustment, $F_{n}$ is increased to $F_{T}$ and $F_{t}$ is decreased, which is precisely the objective of our optimization policy. In the actual grasping, minimizing twist means minimizing $T_{z}$ by changing the gripper configuration to adapt to the object surface. This optimization process is shown in Algorithm \ref{alg:Rigid-Soft}, called Rigid-Soft Interactive Grasping.
    \begin{algorithm}[htbp] 
    \caption{Rigid-Soft Interactive Grasping}
     \label{alg:Rigid-Soft}  
      \begin{algorithmic}[1]  
        \Require  
        Initial Configuration $\theta_i$,  Torque $T_{z,i}$, Decay Rate $\gamma$, Threshold $\delta$, Finger ID $i = 1, 2, 3$.
        \Ensure  
        Optimized Configuration $\hat{\theta}_i$.
        
        \State  Obtain Initial Configuration $\theta_i$, Torque $T_{z,i}$ $ \gets Initial Grasp$
    
        \Function{Grasp Optimization}{$\theta_i, T_{z,i}$}
            \State \bf{parallel for} \textnormal{each finger} \bf{do} 
                \State \qquad  \bf{while} {$T_{z,i} > \delta$} \bf{do}
                    \State \qquad \qquad $\theta_{i} = \theta_{i} - \gamma * T_{z,i} $ 
                \State \qquad \bf{end while}
            \State \bf{end for}
            \State \Return \textnormal{Optimized Configuration} $\hat{\theta}_i$
        \EndFunction
      \end{algorithmic}
    \end{algorithm}

\section{Calibration}
\label{sec:Calibration}
\subsection{Data Collection}
    
\begin{figure}[htbp]
        \begin{centering}
        \textsf{\includegraphics[width=1\columnwidth]{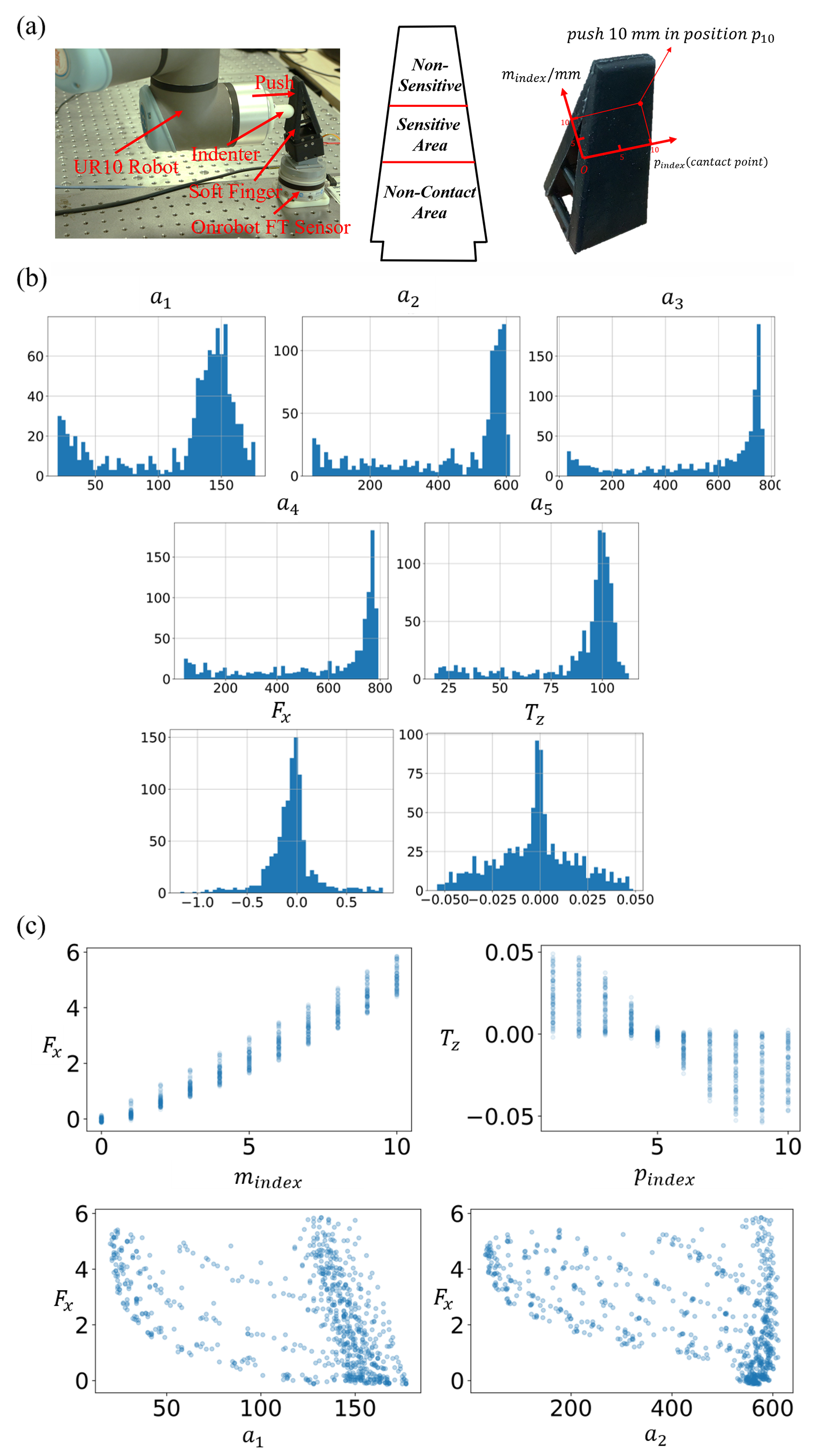} }
        \par\end{centering}
        \caption{Data collection and a summary of collected data. Top: Data collection setup; Bottom: histogram and scatter plots of the calibration experiment data.}
        \label{fig:DataCollection}
    \end{figure}

    We push the soft finger with a 3D-printed indenter in the sensitive area, where the major deformation occurs, as shown in Fig. \ref{fig:DataCollection}. Ten contact points along the horizontal direction were selected, and the feed amount was 10mm at most for each contact point. Contact point $(p_{index})$, feed amount $(m_{index})$, intensity of the transmitted light $(a_{1}, a_{2}, a_{3}, a_{4}, a_{5})$, force and torque information $(F_{n}, T_{z})$ were gathered in this process. Force and torque data were collected from an OnRobot HEX-E v2 mounted on the soft finger's bottom. We collected 1000 data points, and the scatter plots are shown in Fig. \ref{fig:DataCollection}. The force range is from 0 to 6 N, and torque is from -0.05 to 0.05 Nm. In addition to the force and torque prediction, these data could also be interpreted to infer the horizontal direction's contact position. 
    
    Since our tapered fingers have structural asymmetry in the vertical direction, the bend occurs more efficiently when the force is applied to the middle segment of the finger. Under constant normal force, the finger's deformation varies as the contact point moves along the vertical direction. Therefore, we could identify the contact position in the vertical direction under a constant normal force. To achieve this goal, we gathered another set of 100 data points at ten contact points along the vertical direction with the same normal force of $3N$. 

\subsection{Force and Torque}

    To infer the force and torque information, we used the Machine Learning method \cite{scikit-learn} instead of deriving a theoretical model since the latter would be rather tricky due to the complexities of the soft sensor. We chose four widely used regression models, namely Linear model, Decision Tree, Random Forest, and SVM, as our candidates. The dataset of 1000 data points is split into a training dataset of 800 data points and a test dataset of 200 data points. First, we trained the selected four models with the training dataset, respectively. Then, 5-fold cross-validation was used to evaluate the performance and to avoid over-fitting \cite{kohavi1995study}. Finally, the well-trained models were tested on the test dataset. The meta parameters of the models were selected using the gird search toolbox provided by scikit-learning.
    
    We used the commonly used metric, Root Mean Square Error (RMSE), to measure the accuracy of trained models. RMSE is the standard deviation of the residuals, reflecting the model's accuracy. And the equation is 
    \begin{equation}    \label{eq:RMSE}
        RMSE = \sqrt{\frac{1}{m}\sum_{i=1}^{m}(y^{i} - {\hat{y}}^{i})^2}
    \end{equation}
    
    The results are shown in Table \ref{tab:PredictForceTorque}. SVM is the best model for force prediction and the meta parameters are kernel: rbf, C: 1000.0, and gamma: 0.1. The best model for torque prediction is Random Forest, and the meta parameters are bootstrap: False, max features: 2, and $n$ estimators: 100.
    
    \begin{table}[htbp]
        \centering
        \caption{Force and torque prediction results of the optoelectronically innervated tactile finger.}
        \label{tab:PredictForceTorque}
        \resizebox{\hsize}{!}{%
        \begin{threeparttable}
        \begin{tabular}{cccccc}
        \hline
                        & \textbf{RMSE$^*$}             & \textbf{Linear} & \textbf{Tree} & \textbf{Forest} & \textbf{SVM} \\ \hline
        \textbf{}       & \textbf{Baseline}     & 0.534           & 0.004         & 0.082           & 0.166        \\ \cline{2-6} 
        \textbf{Force}  & \textbf{Cross Validation} & 0.541           & 0.331         & 0.225           & 0.196        \\ \cline{2-6} 
        \textbf{}       & \textbf{Test Dataset}     & NA$^{**}$       & NA            & 0.201           & \textbf{0.168}        \\ \hline
        \textbf{}       & \textbf{Baseline}         & 0.004         & 0.000       & 0.001         & 0.021      \\ \cline{2-6} 
        \textbf{Torque} & \textbf{Cross Validation} & 0.004         & 0.004       & 0.003         & 0.021      \\ \cline{2-6} 
        \textbf{}       & \textbf{Test Dataset}     & NA              & NA            & \textbf{0.002}         & 0.021      \\ \hline
    \end{tabular}%

        \begin{tablenotes}
                \footnotesize
                \item[*] Root Mean Square Error (RMSE), whose expression is shown as Eq. (\ref{eq:RMSE}).
                \item[**] Not available, unnecessary.
        \end{tablenotes}
        \end{threeparttable}
        }
    \end{table}

\subsection{Contact Position}

    We used the same dataset and models for predicting the contact position in the horizontal direction as the force prediction. For that in the vertical direction, we used the dataset of 100 samples to train and test the four models. The results in Table \ref{tab:PredictContactPosition} show that random forest models perform best for both directions. The Metaparameters of the best model for horizontal direction are bootstrap: False, max features: 3, $n$ estimators: 100. The Metaparameters for vertical direction are bootstrap: False, max features: 5, $n$ estimators: 3.
    
    \begin{table}[htbp]
        \centering
        \caption{Contact position prediction results of the optoelectronically innervated tactile finger.}
        \label{tab:PredictContactPosition}
        \resizebox{0.48\textwidth}{!}{%
        \begin{tabular}{cccccc}
        \hline
                            & \textbf{RMSE}                  & \textbf{Linear} & \textbf{Tree} & \textbf{Forest} & \textbf{SVM} \\ \hline
        \textbf{}           & \textbf{Baseline}         & 1.574           & 0.000         & 0.246           & 1.581        \\ \cline{2-6} 
        \textbf{Horizontal} & \textbf{Cross Validation} & 1.574           & 0.999         & 0.692           & 0.978        \\ \cline{2-6} 
        \textbf{}           & \textbf{Test}             & NA              & NA            & \textbf{0.569}           & 0.718        \\ \hline
        \textbf{}           & \textbf{Baseline}         & 0.704           & 0.000         & 0.040           & 0.755        \\ \cline{2-6} 
        \textbf{Vertical}   & \textbf{Cross Validation} & 0.761           & 0.250         & 0.099           & 0.190        \\ \cline{2-6} 
        \textbf{}           & \textbf{Test}             & NA              & NA            & \textbf{0.030}          & 0.120        \\ \hline
        \end{tabular}%
        }
    \end{table}

\subsection{Optimal Number of Optical Fibers}

\begin{figure}[htbp]
        \begin{centering}
        \textsf{\includegraphics[width=1\columnwidth]{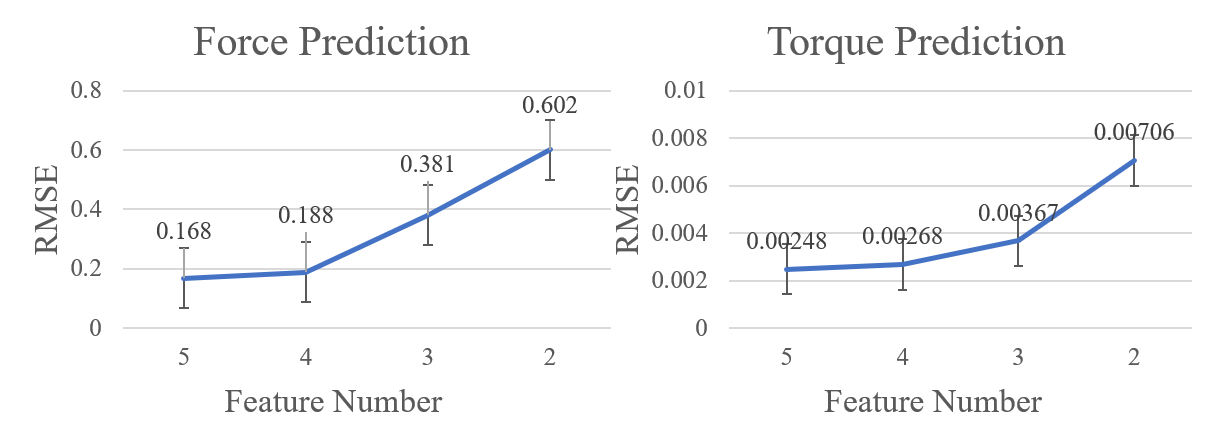}}
        \par\end{centering}
        \caption{Performance prediction as the number of the optic fibers in the soft sensor reduces.}
        \label{fig:FiberCountTest}
    \end{figure}

    To examine the relationship between the number of the optic fibers and the prediction performance, we experimented using only a subset of the five features (intensities of the five optic fibers). We calculated each feature's importance to the final prediction using the forests of trees toolbox and removed the features in a sequence order from less important to more important. Importance of the five features $(a_{1}, a_{2}, a_{3}, a_{4}, a_{5})$ are 0.254, 0.055, 0.199, 0.353, 0.139 for force prediction and 0.359, 0.152 , 0.065, 0.113, 0.309 for torque prediction. The impacts of the feature reduction on the model performance are shown in Fig. \ref{fig:FiberCountTest}.
    
\section{Experiment}
\label{sec:Experiment}
\subsection{Real-time Tactile Sensing \& Prediction}
    
\begin{figure}[t]
        \begin{centering}
        \textsf{\includegraphics[width=1\columnwidth]{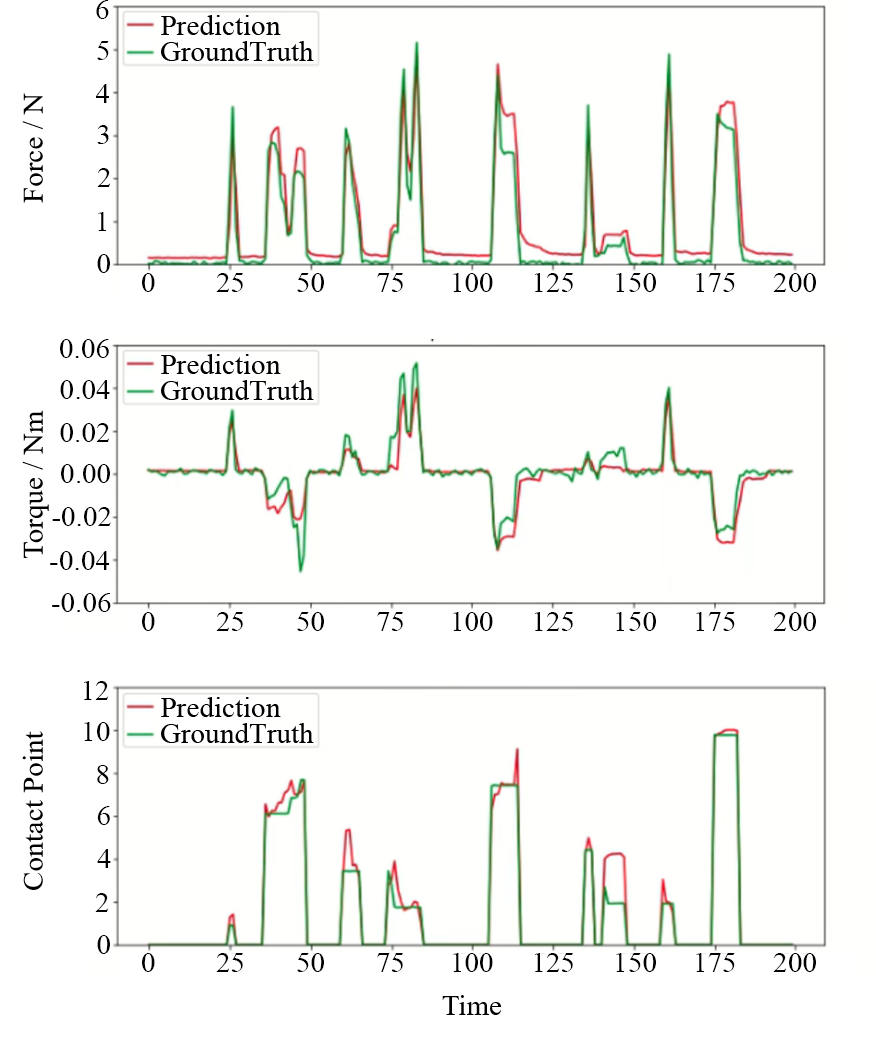} }
        \par\end{centering}
        \caption{Real-time prediction of the force, torque, and contact position. Red line: Prediction value; Green line: Ground truth provided by the OnRobot F/T Sensor.}
        \label{fig:PredictionRealTime}
    \end{figure}

    To test the models' generalization capability, we controlled the robot arm to push the soft finger randomly and predicted the force, torque, and contact position in real-time. Note that the contact point candidates were randomly selected in the continuous range along the horizontal direction instead of 10 discrete points as in the calibration procedure. Similarly, the feed amounts were randomly selected in a continuous range. The results of the real-time prediction are shown in Fig. \ref{fig:PredictionRealTime}.
    
\subsection{Rigid-Soft Interactive Grasping}

\begin{figure}
        \centering
        \includegraphics[width=1\columnwidth]{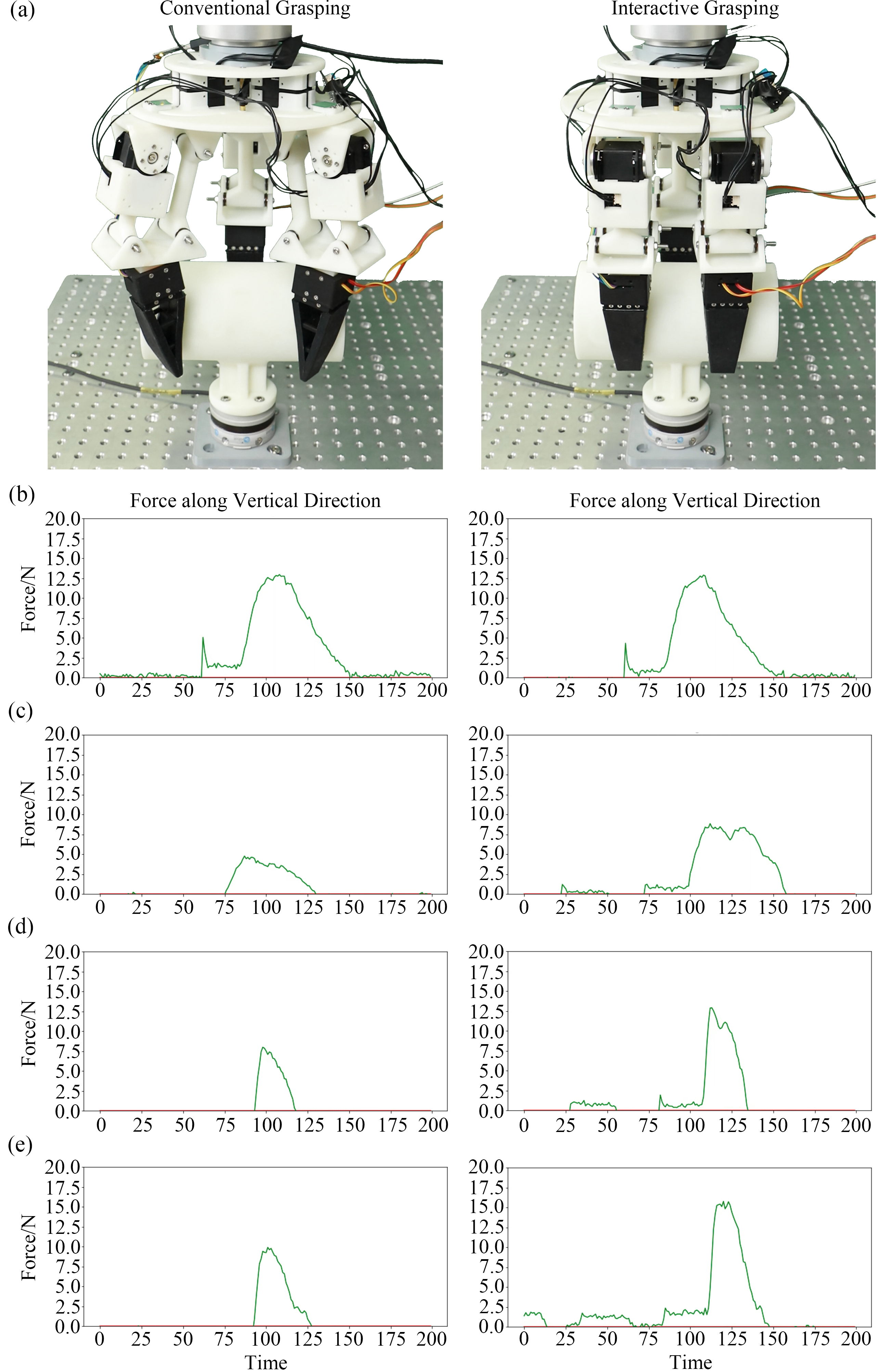}
        \caption{Comparison between the conventional grasping (left) and interactive grasping (right) for four objects. Conventional grasping provides (b) 12.9N for sphere, (c) 4.8N for cube, (d) 8.1N for cuboid, and (e) 9.9N for cylinder. Interactive grasping provides (b) 13.0N for sphere, (c) 9.0N (108\% increase) for cube, (d) 13.1N (61.7\% increase) for cuboid, and (e) 15.8N (59.6\% increase) for cylinder.}
        \label{fig:GraspingComparison}
    \end{figure}

    We applied our soft sensor to a simple grasping process. Three identical soft fingers were fabricated and calibrated. Algorithm \ref{alg:Rigid-Soft} is used in the grasping experiment. First, an initial grasp was conducted with a default circular configuration. The fingers interacted with the target objects, and deformations occurred. The fingers sensed the deformations, which were interpreted to infer force and torque information. The gripper adjusted its configuration with a decay rate of $\gamma$ (10000 is used in this paper) based on the tactile information and adapted to the target object's shape. This adjustment process was repeated until the soft finger's torque, namely the twist of deformation, is less than a threshold (0.002 is used in this paper). Finally, the grasp configuration was optimized, and the robot would try to pick the object up. The target objects were fixed on the table so that we could examine the grasp robustness. The force along the vertical direction while the robot attempted to pick up the object was recorded. A larger value of the vertical force indicates enhanced grasping robustness. The experiment setup is shown in Fig. \ref{fig:GraspingComparison}(a). The conventional grasping with the default circular configuration was also conducted as a benchmark.

    Four objects with standard geometry shapes, namely sphere, cube, cuboid, and cylinder, were used to validate our hypothesis. Gripper was actuated in current-based operation control mode, which means the actuator's torque was constant during the experiment. The results of the comparison between conventional grasping and interactive grasping are shown in Fig. \ref{fig:GraspingComparison}(b)~(e).
    
\section{Discussion}
\label{sec:Discussion}
\subsection{Sensor Calibration for the Proposed Soft Tactile Finger}

     The histogram and scatter plots of the calibration data are shown in Fig. \ref{fig:DataCollection}. From the histogram, we could have an overall look at the dataset, including the ranges and distributions of the calibration data. From the scatter plots, we can see a strong positive linear correlation between feed amount $(m_{index})$ and normal force $(F_{x})$, though it is also influenced by the location of the contact point $(p_{index})$. The sign of torque ($T_{z}$) is decided by whether the contact point is located at the left or right part of the finger. Light intensities $(a_{1}, a_{2}, a_{3}, a_{4}, a_{5})$ have strong nonlinear relations with Force and Torque. In Fig. \ref{fig:DataCollection}, we just plot two of them $(a_{1}, a_{2})$ as examples due to the space limitation. 
    
    Given that the OnRobot Force/Torque sensor has a rough resolution (0.2N for force and 0.001Nm for torque) \cite{OnRobot}, our optoelectronically innervated tactile finger has shown a satisfactory performance on the force and torque predictions with the best model's RMSE less than 0.2N and 0.0025Nm, respectively. For the force prediction, the decision tree model performs best on the baseline, while the cross-validation result shows it is over-fitted. SVM achieves the best performance on the test dataset and is chosen to predict forces in the real grasp experiment. For the torque prediction, the decision tree model also performs best on the baseline but suffers from over-fitting. SVM performs surprisingly bad. Overall the random forest model performs best and is chosen to predict torque in the real grasp experiment.
    
    We treat the contact position as a regression problem despite that the data we collected is discrete. The results in Table \ref{tab:PredictContactPosition} shows our model can generalize on unknown areas. The results of real-time prediction shown in Table \ref{tab:PredictForceTorque} also proved it. When predicting the vertical contact position, we push the soft finger with a constant normal force and get similar horizontal direction performance. It may be used in slippery detection when a gripper grasp using a constant force.

    In the real-time prediction experiment, force, torque, and contact prediction are predicted simultaneously. Fig. \ref{fig:PredictionRealTime} shows that the prediction has a low delay and closely follows ground truth. However, the hysteresis effect is observed when predicting force, probably due to the soft silicone gel's low rebound speed.
    
    Fig. \ref{fig:FiberCountTest} demonstrates the relationship between the number of optic fibers and prediction performance. Generally speaking, the prediction performs worse as the number of optic fibers reduces. In the future, we may further lower the prediction error by increasing the number of optic fibers.
    
\subsection{Towards Rigid-Soft Interactive Grasping}
    
    Generally speaking, the results shown in Fig. \ref{fig:GraspingComparison}(b) indicate a considerable improvement in robustness of grasping using the proposed interactive policy versus conventional open-loop policy, except for the sphere. The reason is that circular configuration is already the most feasible configuration for the sphere. As a result, our interactive policy will not make any difference in grasping robustness. 
    
    However, for cuboid or cylinder, a parallel configuration is preferred. Although the circular configuration could also pick up some light objects, the actual contacting area is very limited since two of the three fingers contact the object with a line. Such grasping is not robust, especially when the object is heavy or some disturbance occurs. The results in Fig.\ref{fig:GraspingComparison}(c)\&(d) show the resistance force is increased by about 60\% (from 8.1N to 13.1N for cuboid and from 9.9N to 15.8N for a cylinder.
    
    The lateral configuration is most feasible for cubic objects since the object is too small to use three fingers. Gripper with two parallel fingers is enough for grasping, and the left one could be used for in-hand manipulation such as pivoting. However, this will involve the force distribution of multi-finger gripper, which is not the paper's scope. In this paper, we focus on single finger optimization and, therefore, improve the final grasp. As a result, the grasping performance is increased by 108\% (from 4.8N to 9.0N). 
    
    In summary, our gripper performs configuration transition between circular configuration, lateral configuration, and parallel configuration according to the target objects' shapes. More than that, our gripper could adapt any irregular objects based on the finger deformation.
    
\section{Final Remarks}
\label{sec:Conclusion}

    In conclusion, this paper presents a novel optoelectronic innervated gripper capable of sensing its soft fingers' deformations, which are further interpreted to infer the force and torque information of each finger using machine learning methods. Also, we propose a grasp optimization policy based on real-time force and torque estimation, closing the loop of grasping control. During the optimization progress, the configuration of the gripper's base is adjusted online. As a result, we could achieve better and more robust grasps that could resist interference.

    This work is a preliminary exploration of an online grasp optimization based on the soft fingers' real-time sensor data. In the future, we would like to investigate further our soft finger sensor's design for better performance, a more in-depth discussion of the optimization progress, and further exploration of multi-sensory information.

\addtolength{\textheight}{-12cm}   


\bibliographystyle{IEEEtran}
\bibliography{reference}

\end{document}